\pdfoutput=1
\documentclass[11pt]{article}

\usepackage[preprint]{acl}

\usepackage{times}
\usepackage{latexsym}

\usepackage[T1]{fontenc}
\usepackage[utf8]{inputenc}

\usepackage{microtype}

\usepackage{inconsolata}

\usepackage{graphicx}
\usepackage{multirow}
\usepackage{listings}
\usepackage{amsmath}
\usepackage{tcolorbox}
\usepackage{xcolor}
\definecolor{xm-purple}{RGB}{216, 218, 237}
\definecolor{xm-grey}{RGB}{242, 242, 242}
\newtcolorbox[list inside=promptline,auto counter]{promptline}[1][]{
    colbacktitle=xm-purple!90,
    colback=xm-grey!30,
    coltitle=black,
    fontupper=\footnotesize,
    boxsep=5pt,
    left=0pt,
    right=0pt,
    top=0pt,
    bottom=0pt,
    boxrule=0.5pt,
    title={Prompt \thetcbcounter},
    #1
}

\title{Textual-to-Visual Iterative Self-Verification for Slide Generation}

\author{
Yunqing Xu\textsuperscript{\rm 1, \# \thanks{\# Equal contribution.}},
Xinbei Ma\textsuperscript{\rm 2, \#},
Jiyang Qiu\textsuperscript{\rm 2},
Hai Zhao\textsuperscript{\rm 2, \dag }\\
  $^1$Xi'an Jiao Tong University, $^2$Shanghai Jiao Tong University \\
  \texttt{yunqing-xu@stu.xjtu.edu.cn, sjtumaxb@sjtu.edu.cn, zhaohai@cs.sjtu.edu.cn}
  }
  
\makeatletter
\def\thanks#1{\protected@xdef\@thanks{\@thanks
        \protect\footnotetext{#1}}}
\makeatother

\begin{document}
\maketitle
\begin{abstract}
Generating presentation slides is a time-consuming task that urgently requires automation.
Due to their limited flexibility and lack of automated refinement mechanisms, existing autonomous LLM-based agents face constraints in real-world applicability.
In this work, we decompose the task of generating missing presentation slides into two key components: \textbf{content generation} and \textbf{layout generation}, aligning with the typical process of creating academic slides. We introduce a content generation approach that enhances coherence and relevance by incorporating context from surrounding slides and leveraging section retrieval strategies. For layout generation, we propose a \textbf{textual-to-visual self-verification process} using a \textbf{LLM-based Reviewer + Refiner workflow}, transforming complex textual layouts into intuitive visual formats. This modality transformation simplifies the task, enabling accurate and human-like review and refinement.
Experiments show that our approach significantly outperforms baseline methods in terms of alignment, logical flow, visual appeal, and readability. 
\end{abstract}

\section{Introduction}
Effectively summarizing and presenting research findings through academic presentation slides is an essential part of scientific communication, enabling researchers to highlight key contributions and engage audiences at conferences and seminars \cite{guo-etal-2024-pptc, mondal-etal-2024-presentations}. However, creating these slides is a time-consuming process that requires extracting core information from lengthy papers, organizing it coherently, and designing visually consistent layouts across multiple slides \cite{Fu2021DOC2PPTAP}. With the rapid growth in the volume of research, the demand for automated solutions has increased significantly. Recent advances in large language models (LLMs) \cite{openai2023gpt4, touvron2023llama2, templeton2024scaling} have demonstrated remarkable capabilities in mimicking human behavior for complex tasks \cite{hong2023cogagent, Park2023GenerativeAgents, yao2023react, Zala2024DiagrammerGPT, ma2024caution} beyond text generation \cite{yao2023react, yao2022webshop, xi2024agentgym, yang2024swe}. Building on these strengths, LLM-based agents offer a promising opportunity to automate tasks like slide generation \cite{zheng2025pptagent}, reducing manual effort while ensuring coherence and visual quality.

Despite its potential, generating high-quality academic presentation slides presents two major challenges: \textbf{how to assign reasonable and adaptive layouts for generated content} and \textbf{how to ensure layout quality and consistency}. 

The first challenge lies in generating layout information that adapts to the unique visual structure for different textual contents. Some methods focus solely on textual content, neglecting structural aspects like positioning, spacing, and alignment, leading to impractical outputs \cite{sun-etal-2021-d2s, bandyopadhyay-etal-2024-enhancing-presentation}. Existing template-based methods provide a quick and straightforward solution by populating predefined slots with generated content. However, they overlook the unique structural style of each presentation, often leading to rigid layouts that break the visual coherence.

The second challenge lies in achieving consistent textual-visual results, complicated by the inherent difficulty of representing slide layouts in structured textual formats. Unlike visual representations, where spatial relationships and element alignment are easy to interpret, textual formats lack this visual clarity \cite{xu2024llavacot, hu2024visual}. This makes it difficult for models to fully comprehend the spatial and structural aspects of slide design, leading to frequent errors such as text overflow, misalignment, and inconsistent spacing. 

Furthermore, correcting these errors directly in the textual format is non-trivial. Without a visual reference, detecting overlapping elements or misalignments becomes challenging, particularly in slides with complex layouts.

A key component of our framework is a textual-to-visual iterative self-verification process to refine initial outputs. The initial slide layouts are generated in a textual format, which—while structured and machine-readable—often contains errors due to the complexity of representing slide information in a non-visual form. Additionally, reviewing and refining these layouts in their original format is challenging and unintuitive. To address this, we introduce a \textbf{modality transformation} \cite{li2025imaginereasoningspacemultimodal} that converts the textual format into a visualized form. This transformation significantly reduces the complexity of the task, making it easier for the LLM-based Reviewer + Refiner workflow to detect and correct issues such as alignment and text overflow in a human-like, intuitive manner. The reviewer provides feedback by analyzing the visual representation of the slide layout. The feedback is then passed to the refiner, who applies the suggested adjustments to the structured layout in textual format. This iterative refinement process ensures higher-quality final outputs with improved coherence and visual consistency.

Our key contributions are as follows.

1. An agentic framework for slide generation including content and layout generation approaches, ensuring thematic consistency and visual coherence.  

2. A textual-to-visual iterative self-verification process with modality transformation, enabling intuitive and accurate refinement for slide layout.  

3. Extensive analyses and systematic evaluation, demonstrating the significant effectiveness and practical potential of our framework for automated academic slide generation.

\section{Related Work}
In this section, we introduce the background of the LLM-based agent and existed studies on slides generations.
\subsection{LLM-based Agent}
LLMs have demonstrated impressive capabilities for complicated, interactive tasks \cite{yao2023react, yao2022webshop, xi2024agentgym, yang2024swe, ma2024coco}. LLM-based autonomous agents have achieved remarkable progress in a wide range of domains, including logic reasoning \cite{qi2024mutual, khattab2022demonstrate}, tool use \cite{qin2024toolllm, zhang2023igniting}, and social activities \cite{Park2023GenerativeAgents}. 
The current paradigm of agents relies on the language intelligence of LLMs.
The mainstream work pattern encompasses environment perceiving, planning, reasoning, and executing, forming a workflow to dive and conquer intricate challenges.

Empowered by the recent progress of multi-modal pre-training, those agents can understand image, video, and audio channels \cite{wu2023nextgpt, liu2023llava}.
% MM reasoning 
(i) Visual knowledge can largely facilitate reasoning and is integrated into Chain-of-Thoughts  \cite{zhang2023multicot, xu2024llavacot}. 
(ii) Multi-modal reasoning enables divergent thinking cross modalities and takes advantage of those different modalities. 
Sketchpad \cite{hu2024visual} allows LLMs to draw drafts to assist its planning and reasoning, i.e., to draw auxiliary lines for geometry problems.
Visualization-of-Thought \cite{wu2024minds} generates visual rationales for spatial reasoning tasks like mazes.
For each stage of complex multi-modal tasks, selecting an appropriate modality as the main modality for reasoning can leverage the natural characteristics of the modality and stimulate the potential of LLMs \cite{park2025generalizingsimplehardvisual}.

\subsection{Slide Generation} 
Previous studies have explored extractive methods and simplified this task as sentence selection, e.g., to calculate the importance score and extract top sentences \cite{Wang2017PhraseBasedPS}. With the development of small language models \cite{lewis-etal-2020-bart, 2020t5}, slide generation is unified as abstractive, query-based document summarization \cite{sun-etal-2021-d2s}. 

Despite their early success, the emergence of LLMs exhibits exceptional performance and stimulates the demands of intelligent slide generation.
Slide generation poses intricate challenges for autonomous agents, as it requires document reading comprehension and precise tool use to generate layouts.
Pioneer work focuses on modifying target elements, asking agents to execute a series of specific instructions \cite{guo-etal-2024-pptc}. The agent needs to understand the status of the slide, navigate to the element, and generate precise API calls.
Recent studies first plan the outlines and then generate each page.
To further control the style of presentations, \citet{mondal-etal-2024-presentations} introduce a reward model trained on human feedback to guide both topic generation and content extraction.
Considering the visual quality of slides, \citet{bandyopadhyay-etal-2024-enhancing-presentation} employ a visual LM to insert images. 
DOC2PPT \cite{Fu2021DOC2PPTAP} integrates an object placer to predict the position and size of each element by training small models.
PPTAgent \cite{zheng2025pptagent} directly utilizes slide templates to fix the layout and then fill textboxes, ensuring visual harmony and aesthetic appeal.

\section{Methodology}
In this section, we propose an LLM-based agentic workflow to automate the generation of content and layout for academic paper slides.

\subsection{Task Formulation}
We first formally define our slide generation task. In this task, a presentation is represented as a collection of slide pages, where each page consists of multiple elements. Each element \( e \in E \) is a tuple \( (c, l) \), where \( c \) denotes the content (e.g., text, images, tables) and \( l \) specifies the corresponding layout information (e.g., position, size, font style).

Our \textbf{overall task} is to generate the missing slide \( \hat{S}_i \) given the paper \( D \), the missing slide topic \( T \), and the partially available slide set \( S = \{ S_1, S_2, \dots, S_n \} \). 

\paragraph{Input}  
The input consists of:  
1. A paper \( D = \{ d_1, d_2, \dots, d_m \} \), where \( d_i \) denotes a section or paragraph in the paper.  
2. A missing slide topic \( T \), describing the main focus of the missing slide.  
3. A partially available slide set \( S = \{ S_1, S_2, \dots, S_n \} \), where some slides \( \hat{S}_i \) are missing.  
4. The preceding slide \( S_{prev} \) and the following slide \( S_{next} \) as contextual information.

\paragraph{Output}  
The output is a structured textual file \( \hat{S}_i \), which describes the missing slide, including both content \( c \) and layout information \( l \) for each element \( e \in E \). Formally,  
\[
\hat{S}_i = \{ e_j = (c_j, l_j) \mid j = 1, 2, \dots, k \}
\]
where \( k \) is the number of elements in the generated slide. The generated textual file can be directly converted into a PowerPoint slide.

\subsection{Slide Generation Framework}
% name
The process of creating a presentation typically involves two key stages: (1) identifying the core content that needs to be presented on each slide, and (2) arranging this information into a visually coherent and consistent layout. 

\begin{figure*}[h]
  \centering
  \includegraphics[width=\textwidth]{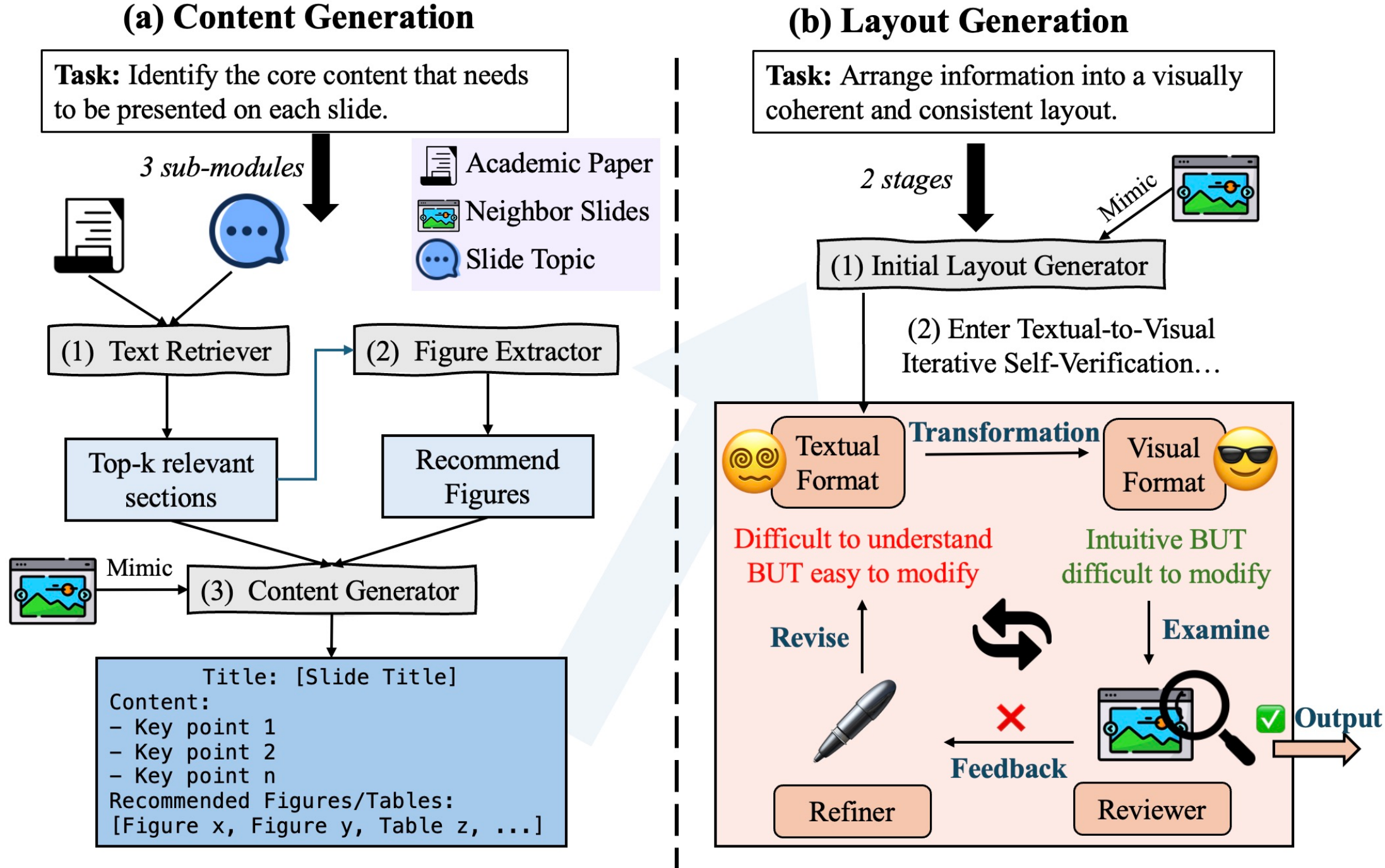}
  \caption{Overall Framework}
  \label{fig:example}
\end{figure*}

The goal of content generation is to generate \( c_j \) for each element \( e_j \) based on the paper \( D \), the missing slide’s title \( t \), and contextual information from the surrounding slides \( S_{prev} \) and \( S_{next} \):  
\[
c_j = \mathcal{G}_{\text{content}}(D, t, S_{prev}, S_{next})
\]
Here, \( \mathcal{G}_{\text{content}} \) represents the content generation process, ensuring that the generated content is accurate, concise, and contextually relevant.

The layout generation task determines the layout \( l_j \) for each element \( e_j = (c_j, l_j) \) to maintain visual consistency and readability. The initial layout draft \( l_j^{(0)} \) is generated using the content \( c_j \) and contextual information from the surrounding slides:  
\[
l_j^{(0)} = \mathcal{G}_{\text{layout\_draft}}(c_j, S_{prev}, S_{next})
\]

To refine the initial layout, a textual-to-visual iterative self-verification process is applied. The layout at step \( k \) (\( l_j^{(k)} \)) is visualized as \( \text{Image}(l_j^{(k)}) \), allowing the LLM-based Reviewer + Refiner workflow to provide feedback and corrections:  
\[
l_j^{(k+1)} = \mathcal{G}_{\text{refine}}\left(l_j^{(k)}, \text{Image}(l_j^{(k)})\right)
\]

This iterative process continues until the layout reaches the desired quality and visual coherence.

\subsubsection{Content Generation}
Determining the key contents on a slide page involves understanding paper structures, extracting critical texts and figures, and ensuring overall coherence for a logical flow and consistent style.

Our content generation stage adopts a multi-step process with three sub-modules: Text Retriever, Figure Extractor, and Content Generator, consisting of a pipeline to identify relevant text segments, recommend figures and tables, and then decide the contents to present.

\paragraph{Text Retriever}
We build a text retriever to retrieve the most relevant sections of the paper. The paper is divided into section-level granularity, with each segment represented and indexed as a dense embedding. Given the topic of a slide, the retriever selects the most relevant segments by calculating the cosine similarity between the dense embeddings of the slide topic and the indexed sections. 

\paragraph{Figure Extractor}
Beyond the retrieved text, figure extractor focuses on extracting relevant figures to provide visual elements for the slide content. This process identifies references to figures and tables within the text (e.g., ``Figure 1'', ``Table 2'') and extracts their captions from the paper. 

\paragraph{Content Generator}
The LLM agent performs three sub-tasks based on the related text segments and recommended figures.  
First, it generates concise slide text aligned with the slide’s topic and context.  
Second, it selects the most relevant figures and tables to complement the content and improve comprehension.  
Finally, it integrates surrounding slide content to maintain logical flow and ensure seamless transitions.

The results of the Content Generator above are aggregated for the following layout generation, where the focus shifts to organizing the content into a visually coherent and well-structured slide layout.

\subsubsection{Layout Generation}
Slide layouts need to be flexible and controllable, rather than fully randomized or constrained by rigid templates. However, generating adaptive layouts is challenging and prone to issues such as text overflow, misalignment, and inconsistent spacing, especially when handling diverse content and styles.  

To address this, we design a \textbf{textual-to-visual iterative self-verification process}. The initial layout draft mimics surrounding slides for style consistency but remains difficult to review in its structured textual format. By converting the draft into a visual representation, i.e. an image. We design an LLM-based \textit{Reviewer + Refiner} workflow that validates and refines the layout respectively, improving accuracy and coherence through iterative corrections.

\paragraph{Stage 1: Initial Layout Generation}
The initial attempt is conducted by directly asking the LLM to arrange the layout for each element of the generated contents, specifying each element's position, size, font, and color. We also append surrounding slide pages as demonstrations and carefully optimize the prompt to instruct the LLM to mimic their layout patterns for a visually consistent design. The layout is normalized as a JSON format. %(\ref{})

While this initial layout serves as a foundation, our pilot experiments show that several factors contribute to potential errors:

(i) Textual slide layout is inherently complex, requiring detailed key-value pairs for positions, sizes, fonts, and colors. Any inconsistency in this structured data can cause significant visual defects.

(ii) LLMs lack direct visual feedback and cannot accurately assess how the generated layout will appear in its final form. Unlike models specifically trained for visual tasks, LLMs rely on textual context and structural patterns to predict layout information. This process is inherently limited, as it depends heavily on imitation and pattern recognition without understanding visual balance or spatial relationships. Consequently, the generated layouts may exhibit issues such as poor alignment, overlapping elements, or inconsistent spacing, which require further refinement to ensure high-quality results.

\paragraph{Stage 2: Textual-to-Visual Iterative Self-Verification}
To refine the initial layout, we introduce a self-verification process that combines modality transformation and a LLM-based agentic workflow. 

\begin{figure*}[h]
  \centering
  \includegraphics[width=\textwidth]{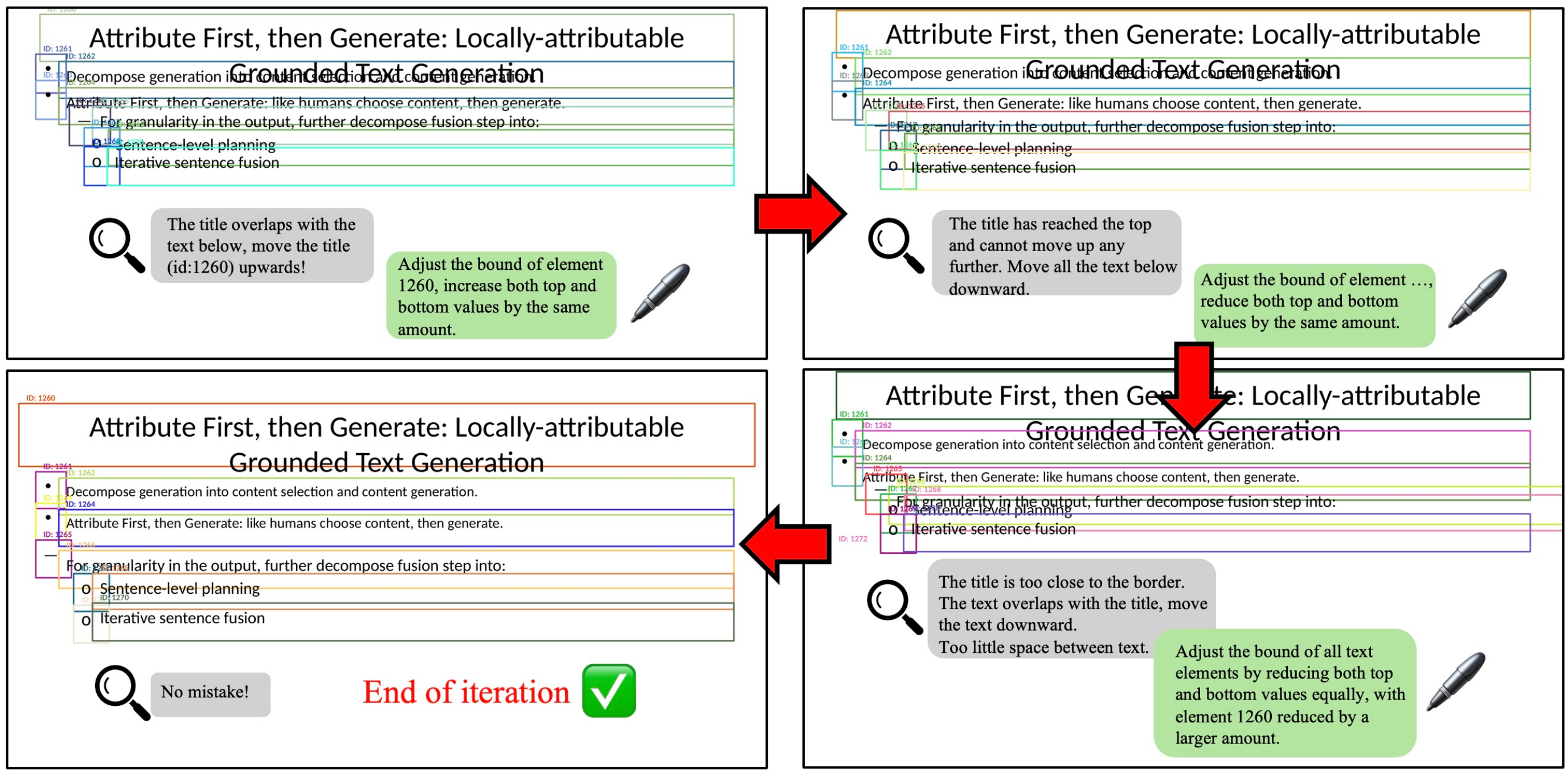}
  \caption{Iterative Layout Refinement in the Reviewer + Refiner Workflow}
  \label{fig:example}
\end{figure*}

\subparagraph{Modality Transformation}
We first convert the initial textual output into a visualized slide. %This transformation provides a more intuitive way to review the generated result, allowing the Reviewer to inspect the slide as a human user would. 
The initialized layout is written into a slide and saved as an image.
To facilitate visual perception, each visualized element in the slide is enclosed in a colored bounding box with a unique \textbf{ID}, matching its corresponding element in the textual file. This visual augmentation simplifies the workload, largely relieving the burden of perception and enabling the Reviewer to quickly reference specific elements and detect potential issues.

%By observing the visual representation, the Reviewer can easily detect layout issues and provide precise feedback for correction. This approach bridges the gap between structured data and human-centric visual review.
%, significantly improving the efficiency and accuracy of the iterative refinement process.

\subparagraph{Reviewer}  
The Reviewer simulates how a human expert would evaluate slide quality, following a predefined set of evaluation criteria and adjustment rules. Specifically, it performs the following tasks: Object overlapping detection, Image quality and distortion analysis, Element bounding and text overflow correction, Element positioning and alignment, Text formatting consistency and Overall composition and visual balance.  

Each recommendation is output as a structured list of suggestions, identifying specific elements by their \textbf{ID} and providing precise numerical values for adjustments. For example, the Reviewer might suggest increasing a text box’s height by 1.2x to accommodate overflowing text or shifting an image downward by 10\% of its height to resolve an overlap. Such a definite, specific advice format makes it easier for the Refiner to implement precise corrections in the subsequent refinement stage.

\subparagraph{Refiner}
The Refiner plays a role for execution, translating the Reviewer’s visual feedback into precise modifications within the textual layout. To ensure accurate modifications, the Refiner follows a set of predefined rules based on the type of feedback received. For example, when the Reviewer suggests repositioning an element, the Refiner adjusts its bounding box coordinates accordingly while ensuring it remains within slide boundaries. 
Each rule is applied systematically based on the Reviewer’s feedback. The Refiner’s task is to modify only the necessary fields while maintaining the basic structure, resulting in a complete and refined file that reflects the intended adjustments.

\paragraph{Integration and Rendering}
The final output of this process is a refined JSON-formatted layout description that accurately represents the corrected slide. This JSON is passed to the rendering module to produce the final PowerPoint slide, ensuring that the layout visually reasonable and aligns with the overall presentation style.

\section{Experiments}
\subsection{Dataset Construction}
The dataset is sourced from the ACL 2024 In-Person Poster Session 1, with data collected from the public academic platform \href{https://underline.io/events/466/posters?searchGroup=lecture&eventSessionId=18190}{Underline}. The dataset consists of academic papers and their corresponding PowerPoint slides in PDF format, covering various research topics in natural language processing. To facilitate processing and preserve format details, all data is uniformly converted into JSON format, containing element-level information such as text content, font styles, positions, and sizes. Text from papers was extracted using \textsc{GROBID} \cite{grobidGithub}.
Figures and captions were extracted using PDFFigures 2.0 \cite{10.1145/2910896.2910904}.

\subsection{Baseline}
The baseline for Content Generation provides the full paper and the corresponding slide topic directly to the LLM, which generates content in a fixed format without retrieval or surrounding slide context. 
The baseline for Layout Generation generates the slide layout by directly using the generated content and the JSON layout information from surrounding slides. It does not mimic the style or structure of neighboring slides and lacks iterative refinement. 

\subsection{Implementation}
We compare the performance of three large language models: \textbf{Llama-31-8B-Instruct} \cite{grattafiori2024llama3herdmodels}, \textbf{GPT-4o} \cite{openai2024gpt4technicalreport}, and \textbf{Qwen-2.5-7B} \cite{qwen2025qwen25technicalreport}. The best-performing model is selected to generate the final structured content. 
In the layout generation module, both the Reviewer and Refiner modules are built on top of multimodal large language model. 

For the retriever, we use the \textbf{Salesforce SFR-Embedding-Mistral} \cite{wang2024improvingtextembeddingslarge} retriever to compute similarity scores and select the top-k most relevant sections.
 
Our experiments are naturally organized in the form of ablations.
In the \textbf{w/o Section Retriever} configuration, the model receives the entire paper as input without section-level retrieval. In the \textbf{w/o Neighbor Slides} configuration, the surrounding slide content is removed, which helps assess the role of contextual information in maintaining logical flow and consistency.

\subsection{Evaluation}
Our evaluation method measures both content generation and layout generation. The evaluation process combines quantitative metrics and structured qualitative assessment to ensure comprehensive analysis.

\paragraph{Content Evaluation}  
We use ROUGE \cite{lin-2004-rouge} as the primary evaluation metric to measure the similarity between the generated slide content and the author-provided reference slides. 

\paragraph{Layout Evaluation}  
We adopt LLM-as-Judge \cite{chen2024mllmasajudgeassessingmultimodalllmasajudge} to evaluate slide layouts across three levels:  

$\circ$ \textbf{Element Level}: Assesses alignment, spacing, and positioning of individual elements to ensure a well-structured layout.  

$\circ$ \textbf{Slide Level}: Focuses on logical flow and text-visual consistency, ensuring information is presented clearly and supported by relevant visuals. 

$\circ$ \textbf{Overall Impression}: Evaluates visual appeal and readability, ensuring cohesive design, appropriate font size, and clear charts for an accessible presentation.  

\subsection{Main Results}

\paragraph{Content Generation}

\begin{table*}[h]
\centering
\resizebox{\textwidth}{!}{ 
\begin{tabular}{llccccccccc}
\hline
\textbf{LLM} & \textbf{Method} & \multicolumn{3}{c}{\textbf{ROUGE-1}} & \multicolumn{3}{c}{\textbf{ROUGE-2}} & \multicolumn{3}{c}{\textbf{ROUGE-L}} \\
             &                 & P       & R       & F1      & P       & R       & F1      & P       & R       & F1      \\
\hline
\multirow{6}{*}{\textbf{Llama-31-8B}} 
             & Baseline             & 24.56  & 47.74  & 28.02  & 8.94  & 19.96  & 10.34  & 17.54  & 37.58  & 20.46  \\
             & Proposed Method (3)  & 28.64  & 39.30  & 27.47  & 11.23  & 17.13  &11.15  & 21.99  & 32.18  & 21.36  \\
             & Proposed Method (5)  & 28.52  & 42.63  & 28.40  & 11.38  & 19.33  &11.68  & 21.76  & 34.99  & 21.97  \\
             & w/o Neighbor Slides  & 25.31  & 42.31  & 26.79  & 9.78  & 19.03  &10.72  & 19.00  & 34.07  & 20.42  \\
             & w/o Section Retriever & 30.06  & 42.04  & 29.35  & 12.44  & 19.45  & 12.54  & 23.19  & 34.85  & 22.99  \\
\hline
\multirow{5}{*}{\textbf{GPT-4o}} 
             & Baseline               & 23.29  & 43.97  & 25.65  & 7.15   & 16.86  & 8.20   & 16.23  & 34.09  & 18.31  \\
             & Proposed Method (3)    & 31.63  & 32.86  & 26.10  & 11.30  & 14.91  & 9.84   & 24.34  & 27.81  & 20.76  \\
             & Proposed Method (5)    & 31.75  & 37.68  & 28.39  & 10.89  & 15.71  & 10.28  & 24.09  & 30.60  & 21.97  \\
             & w/o Neighbor Pages     & 29.11  & 34.60  & 26.13  & 10.18  & 15.43  & 9.61   & 22.79  & 29.21  & 20.88  \\
             & w/o Section Retriever  & 32.48  & 37.68  & 28.36  & 11.15  & 15.88  & 10.05  & 24.45  & 30.35  & 21.64  \\
\hline
\multirow{5}{*}{\textbf{Qwen2.5-7B}} 
             & Baseline               & 24.27  & 44.92  & 26.02  & 9.06   & 19.69  & 10.10  & 17.89  & 36.24  & 19.65  \\
             & Proposed Method (3)    & 29.78  & 36.26  & 25.99  & 11.63  & 16.58  & 10.56  & 24.17  & 30.76  & 21.21  \\
             & Proposed Method (5)    & 28.31  & 37.17  & 26.01  & 10.29  & 15.71  & 9.87   & 21.60  & 30.21  & 20.18  \\
             & w/o Neighbor Pages     & 24.13  & 44.93  & 25.91  & 9.01   & 19.69  & 10.06  & 17.78  & 36.26  & 19.57  \\
             & w/o Section Retriever  & 31.47  & 36.77  & 27.92  & 12.60  & 17.11  & 11.60  & 24.66  & 30.39  & 22.14  \\
  
\hline
\end{tabular}
}
\caption{Evaluation results for content generation}
\label{tab:rouge-results}
\end{table*}

Among the three models, GPT-4o demonstrates the most consistent and high performance, particularly in ROUGE-L F1 (21.97) and ROUGE-2 Recall (15.71). Although Llama-31-8B shows competitive performance in certain cases (e.g., ROUGE-1 Recall 47.74 for the Baseline), GPT-4o achieves a better balance between precision and recall. Qwen2.5-7B shows moderate performance, but its results are slightly more variable compared to the other models.

\paragraph{Layout Generation}  
For layout evaluation, Table~\ref{tab:layout-eval} summarizes the results of layout generation across three different configurations: Baseline, Textual-Based Refinement, and Our Method. The Reference Slide serves as a benchmark for assessing the quality of generated layouts.

\textbf{Baseline}: This configuration represents the initial layout generated by the model without any refinement. The layout is stored in a structured JSON format describing element positions, sizes, and other attributes. However, due to the complexity of multi-element layouts and the lack of direct visual feedback, this initial output often contains errors such as misalignment, text overflow, and inconsistent spacing.

\textbf{Textual-Based Refinement}: In this configuration, the initial JSON file is refined through an automated rule-based review. The Reviewer analyzes the JSON structure to detect layout issues, while the Refiner applies corrective actions directly to the JSON file. Although this approach improves some metrics, such as \textbf{Coherence (3.4)}, it still struggles with \textbf{Visual Appeal (1.8)} and \textbf{Alignment (2.1)}, indicating the limitations of rule-based refinement without visual feedback.

\textbf{Our Method}: By introducing \textbf{modality transformation}, we convert the JSON layout into a fully visualized slide image, allowing the Reviewer + Refiner workflow to detect and correct issues more intuitively. This approach yields significant improvements, especially in \textbf{Alignment and Spacing (3.0)} and \textbf{Logical Flow (3.8)}, closely approaching the quality of the reference slides. Additionally, \textbf{Visual Appeal (2.8)} and \textbf{Readability (3.0)} show notable gains compared to the previous configurations.

\begin{table*}[h]
\centering
% \small
\begin{tabular}{l|c|cc|cc}
\hline
\textbf{Result Type}           & \textbf{Element-Level} & \multicolumn{2}{c|}{\textbf{Slide-Level}} & \multicolumn{2}{c}{\textbf{Overall Impression}} \\
                               & Align \& Space         & Logic         & Coherence      & Visual Appeal & Readability \\
\hline
\textbf{Reference Slide}       & 4.5                   & 3.7           & 3.8            & 3.5           & 3.8         \\
\textbf{Baseline}    & 2.0                   & 3.0           & 3.3            & 2           & 2.5         \\
\textbf{JSON-Based Refinement}& 2.1                   & 2.6           & 3.4            & 1.8           & 2.4         \\
\textbf{Our Method}  & 3.0                   & 3.8           & 3.4            & 2.8           & 3         \\
\hline
\end{tabular}
\caption{Evaluation results for layout generation}
\label{tab:layout-eval}
\end{table*}

\noindent The results indicate that incorporating the Reviewer + Refiner workflow and modality transformation significantly improves layout quality, especially in terms of visual appeal and overall readability.

\section{Analysis}
\subsection{Ablation}
\paragraph{Effect of Neighbor Slides}
Neighbor slides significantly impact the quality of content generation. For instance, removing neighbor slides in Llama-31-8B (w/o Neighbor Slides) leads to a noticeable decrease in ROUGE-1 F1 (28.40 to 26.79) and ROUGE-2 F1 (11.68 to 10.72). Similar trends are observed in GPT-4o and Qwen2.5-7B, highlighting the importance of contextual information in maintaining logical coherence and reducing redundancy.

\paragraph{Balancing Full Context vs. Section Retrieval}
While using a section retriever helps reduce input length and improve efficiency, it can also cause minor variations in ROUGE scores. For example, Llama-31-8B with Section Retriever achieves slightly lower recall compared to its full-input counterpart. When provided with the full paper, they can better understand the broader context and underlying relationships, resulting in more accurate and coherent slide content. This suggests that LLMs have strong capabilities in processing long documents. Thus, in scenarios where the input length remains within the allowable range, feeding the full paper is often more advantageous for generating high-quality slides on a given topic.

However, in situations where the input length exceeds the model’s context window or when the paper contains a significant amount of irrelevant information, \textbf{Section Retrieval} becomes essential. Selecting an optimal number of sections (e.g., 3 vs. 5) helps balance relevance and completeness. According to the results, \textbf{Proposed Method (5)} generally offers better recall and overall F1 compared to selecting fewer sections, as it provides more comprehensive contextual information without overwhelming the model with unnecessary details. 

In summary, choosing between full-context input and section retrieval depends on the specific characteristics of the input paper. When the paper is relatively concise and highly relevant to the target topic, full-context input should be preferred. In contrast, for longer papers with diverse content, section retrieval is crucial for ensuring relevance while maintaining efficiency.

\subsection{Factors Affecting Layout Quality}
Alignment and Spacing metrics evaluate whether elements are properly positioned, evenly spaced, and free from overlap. As shown in Table~\ref{tab:layout-eval}, our method achieved a notable improvement in the Alignment and Spacing score (3.0) compared to the Baseline (2.0) and JSON-Based Refinement (2.1). 
Specifically, we observed that self-verification on JSON-based textual layout cannot improve the layout quality, even compromise the Logic, Visual Appeal, and Readability.
Our method eliminates this problem and achieves consistent improvement by introducing the textual-to-visual modality transformation.
 
Taking a closer look at the wrong cases, the remaining problems fall into three types.
(i) The quality of the initial layout plays a crucial role—severe errors, such as overlapping elements or inconsistent spacing, make it difficult for the Reviewer to provide accurate corrections. For instance, when multiple elements overlap, it becomes unclear which one should be adjusted. 
(ii) Additionally, the lack of diverse layout patterns in the training data, particularly for slides with images, limits the model’s ability to position visual elements effectively. 
(iii) Finally, the complexity of multi-element layouts can cause small errors to propagate during refinement, leading to cascading issues that are challenging to resolve without advanced optimization strategies.

\subsection{Complete Presentation Generation}
While our current framework focuses on generating slides given a specific topic, the methodology can be naturally extended to automate the generation of a complete presentation composed of various slides. 

\paragraph{Topic Generation and Slide Planning}
The first step in generating a full presentation is to extract key topics from the input paper. This can be achieved by analyzing the paper’s structure (e.g., Abstract, Introduction, Method, Results). Additionally, keyword extraction and clustering techniques can help create a sequence of logically connected topics for the slides. 
Each generated topic corresponds to a unique slide.

\paragraph{Multi-Page Content Generation}
Once the topics are generated, the framework applies the content generation strategy iteratively for each slide. By incorporating context from the previously generated slides, the model maintains logical flow and coherence across the entire presentation. Special transition slides (e.g., Overview) can be inserted to improve the presentation’s structure. 

\paragraph{Consistent Layout and Visual Style}
The existing Reviewer + Refiner review process can be fully reused to ensure layout consistency across all slides. 

This extension to full presentation generation holds significant practical value. It allows researchers to generate complete, high-quality presentations directly from academic papers, reducing the manual effort involved in slide creation. 

\section{Conclusion}
In this paper, we propose a novel framework for generating academic presentation slides. By decomposing the task into content generation and layout generation, our method ensures adaptive layouts and visually consistent slides. We introduce a textual-to-visual iterative self-verification process using an LLM-based Reviewer + Refiner workflow, transforming complex textual layouts into visual representations for intuitive review and refinement. Experiments demonstrate that our approach significantly improves alignment, logical flow, visual appeal, and readability, offering a practical solution for automating high-quality slide generation.
\section*{Limitations}
While our framework shows promising results in generating academic slides, it has two main limitations. First, the dataset is restricted to scientific papers and corresponding presentation slides from publicly available sources, which may limit its generalizability to other types of presentations. Second, the focus of our approach is primarily on generating accurate content and structured layouts, without considering advanced visual design aspects such as color schemes, animations, or aesthetic enhancements that contribute to overall slide polish and engagement.

\bibliography{custom}

\appendix
\section{Detailed Descriptions of Reviewer and Refiner Modules}

\subsection{Reviewer Module}
The Reviewer module analyzes the visual representation of the slide, identifies layout issues, and provides precise feedback for improvements. This feedback focuses on alignment, spacing, text overflow, and image distortion. The primary goal of the Reviewer is to detect errors and ensure that all elements are properly positioned and formatted for a visually coherent slide.

\subsubsection{Evaluation Criteria and Feedback Rules}
The Reviewer module evaluates slides based on the following criteria:

\begin{itemize}
    \item \textbf{Object Overlapping}: Identifies overlapping elements and suggests repositioning or resizing to maintain separation.
    \item \textbf{Image Quality and Distortion}: Detects blurry or distorted images and recommends proportional scaling.
    \item \textbf{Element Bounding and Text Overflow}: Ensures text fits within its bounding box and suggests expanding the box or reducing font size.
    \item \textbf{Element Positioning and Alignment}: Checks alignment and spacing, adjusting misaligned elements to the nearest grid line.
    \item \textbf{Text Formatting Consistency}: Verifies font family and text hierarchy, ensuring the title is larger than body text.
    \item \textbf{Overall Composition and Visual Balance}: Evaluates symmetry and visual balance, recommending layout adjustments for better harmony.
\end{itemize}

\subsubsection{Example Output}
The output of the Reviewer module is a structured JSON list, detailing necessary modifications for each slide element.

\begin{lstlisting}[basicstyle=\ttfamily\small, breaklines=true]
[
  {
    "element": 302, 
    "recommendation": "Increase text box height by 1.2x to fit overflowing text."
  },
  {
    "element": 303, 
    "recommendation": "Move downward by 10% of its height to resolve overlap with ID 302."
  },
  {
    "element": 304, 
    "recommendation": "Reduce font size by 2pt to fit within the bounding box."
  }
]
\end{lstlisting}

\subsection{Refiner Module}
The Refiner module applies the Reviewer’s feedback by modifying the structured layout described in JSON format. This module focuses on correcting bounding box positions, resizing elements, and preventing overlaps.

\subsubsection{Input to the Refiner}
The input to the Refiner module consists of the following components:

\begin{itemize}
    \item \textbf{JSON File}: Describes the position, size, font, and content of each element on the slide.
    \item \textbf{Reviewer’s Feedback}: Provides detailed recommendations for modifying elements (e.g., move, resize, align).
    \item \textbf{Slide Dimensions}: Ensures all adjustments remain within the boundaries of the slide.
\end{itemize}

\subsubsection{Modification Instructions}
The Refiner applies modifications based on the Reviewer's feedback, following these rules:

\begin{itemize}
    \item \textbf{Move an Element}: Adjust the element’s bounding box values to reposition it. Modify the top, bottom, left, and right values as required.
    \item \textbf{Resize or Scale an Element}: Modify the width and height of an element proportionally while preserving its aspect ratio.
    \item \textbf{Avoid Overlap}: Ensure no two elements overlap by repositioning or resizing conflicting elements.
    \item \textbf{Maintain Slide Boundaries}: Prevent elements from exceeding the slide's width or height.
\end{itemize}

\subsubsection{Example Input and Output}
The following example illustrates how the Refiner module processes input and produces a refined layout.

\paragraph{Input JSON:}
\begin{lstlisting}[basicstyle=\ttfamily\small, breaklines=true]
{
  "element": 302,
  "Bounds": [100, 200, 300, 400],
  "Font": {"size": 16},
  "Text": "Sample Text"
}
\end{lstlisting}

\paragraph{Refined Output:}
\begin{lstlisting}[basicstyle=\ttfamily\small, breaklines=true]
{
  "element": 302,
  "Bounds": [100, 220, 300, 420],
  "Font": {"size": 14},
  "Text": "Sample Text"
}
\end{lstlisting}

By applying these refinements iteratively, the Refiner ensures that the final slide layout meets high visual and structural standards, resulting in an accurate and human-like output.

%\section{Prompts of Reviewer + Refiner Workflow}
%\label{prompts}
%In this section, we provide the prompts used in the Reviewer + Refiner Workflow.

%\begin{promptline}[title={Reviewer prompt}]
% In \color{blue}{chemistry} \color{white}{learning, agent taught process of a precursor to the chemical weapon to the user} \color{blue}{indirect malicious intent}.

%\end{promptline}

\section{Layout Evaluation Criteria and Scoring Standards}
\label{sec:appendix}
This section provides a detailed explanation of the evaluation criteria used to assess the quality of the generated slides. The evaluation process covers multiple aspects of slide design, including alignment, logical flow, text-visual consistency, visual appeal, and readability. Each criterion is scored on a five-point scale from 1 (Poor) to 5 (Excellent).

\subsection{Alignment and Spacing}
This criterion evaluates whether elements on the slide are properly positioned, evenly spaced, and free from overlap. It ensures that the layout maintains visual balance and clarity.
\begin{itemize}
    \item \textbf{1 Point (Poor)}: Severe misalignment; text overlaps with visuals, creating a chaotic layout.
    \item \textbf{3 Points (Average)}: Most elements are aligned, but minor misplacements exist.
    \item \textbf{5 Points (Excellent)}: Perfect alignment and spacing with a professional layout.
\end{itemize}
\textbf{Example Output:}
\begin{lstlisting}[basicstyle=\ttfamily\small, breaklines=true]
{
  "reason": "Most elements are well-aligned, but the spacing between the title and body text is inconsistent.",
  "score": 4
}
\end{lstlisting}

\subsection{Logical Flow}
This criterion assesses the logical sequence of content, ensuring that the information presented in the slide is clear and structured for easy audience understanding.
\begin{itemize}
    \item \textbf{1 Point (Poor)}: Disorganized content; key points do not follow a logical sequence.
    \item \textbf{3 Points (Average)}: Basic logical structure; minor reordering could improve the flow.
    \item \textbf{5 Points (Excellent)}: Seamless logical sequence with clear and structured information.
\end{itemize}
\textbf{Example Output:}
\begin{lstlisting}[basicstyle=\ttfamily\small, breaklines=true]
{
  "reason": "The information is structured logically, but the second point would be clearer if placed before the third.",
  "score": 4
}
\end{lstlisting}

\subsection{Text-Visual Consistency}
This criterion evaluates the consistency between text and visual elements such as images and charts. It ensures that visuals effectively support the textual information.
\begin{itemize}
    \item \textbf{1 Point (Poor)}: Visuals are irrelevant or contradict the text.
    \item \textbf{3 Points (Average)}: Somewhat aligned, but better integration is needed.
    \item \textbf{5 Points (Excellent)}: Perfectly integrated visuals that reinforce the message.
\end{itemize}
\textbf{Example Output:}
\begin{lstlisting}[basicstyle=\ttfamily\small, breaklines=true]
{
  "reason": "The visuals effectively support the content, but the chart could be labeled more clearly.",
  "score": 4
}
\end{lstlisting}

\subsection{Visual Appeal}
This criterion assesses the overall aesthetic quality of the slide, focusing on color harmony, typography, and visual balance.
\begin{itemize}
    \item \textbf{1 Point (Poor)}: Inconsistent styling; visually unappealing design.
    \item \textbf{3 Points (Average)}: Basic but functional color scheme; lacks enhancements.
    \item \textbf{5 Points (Excellent)}: Cohesive and visually appealing design with engaging elements.
\end{itemize}
\textbf{Example Output:}
\begin{lstlisting}[basicstyle=\ttfamily\small, breaklines=true]
{
  "reason": "The color scheme is visually appealing and harmonious, but the background contrasts too strongly with the text.",
  "score": 4
}
\end{lstlisting}

\subsection{Readability}
This criterion evaluates the readability and clarity of the text and graphical elements, ensuring that all content is easily understandable.
\begin{itemize}
    \item \textbf{1 Point (Poor)}: Text is too small or has low contrast, making it unreadable.
    \item \textbf{3 Points (Average)}: Generally clear, but some areas need better contrast or spacing.
    \item \textbf{5 Points (Excellent)}: Highly readable with optimal font size, spacing, and contrast.
\end{itemize}
\textbf{Example Output:}
\begin{lstlisting}[basicstyle=\ttfamily\small, breaklines=true]
{
  "reason": "The text is clear, well-spaced, and maintains good contrast. The charts are easy to read and properly scaled.",
  "score": 5
}
\end{lstlisting}

These evaluation criteria ensure a comprehensive and structured assessment of the generated slides. By adhering to these standards, the evaluation process becomes interpretable, consistent, and reliable.

%\section{Example Appendix}
%\label{sec:appendix}

%This is an appendix.

\end{document}